\documentclass[a4paper,fleqn]{cas-sc}

\usepackage[numbers]{natbib}

\def\tsc#1{\csdef{#1}{\textsc{\lowercase{#1}}\xspace}}
\tsc{WGM}
\tsc{QE}
\tsc{EP}
\tsc{PMS}
\tsc{BEC}
\tsc{DE}

\usepackage{graphicx}%
\usepackage{multirow}%
\usepackage{amsmath,amssymb,amsfonts}%
\usepackage{amsthm}%
\usepackage{mathrsfs}%
\usepackage[title]{appendix}%
\usepackage{xcolor}%
\usepackage{textcomp}%
\usepackage{manyfoot}%
\usepackage{booktabs}%
\usepackage{algorithm}%
\usepackage{algpseudocode}%
\usepackage{listings}%
\usepackage{enumitem}
\usepackage{lipsum}
\usepackage{subcaption}
\usepackage{mwe}
\usepackage{adjustbox}
\usepackage{fancyhdr}
\usepackage{caption}
\usepackage{csquotes}
\usepackage{colortbl}
\usepackage{comment}
\usepackage{placeins}
\usepackage{float}

\theoremstyle{thmstyleone}%
\theoremstyle{thmstyletwo}%

\theoremstyle{thmstylethree}%

\begin{document}
\let\WriteBookmarks\relax
\def\floatpagepagefraction{1}
\def\textpagefraction{.001}

\shorttitle{Forward-Only Convolutional Neural Networks with Learnable Channel–Class Assignment}    

\shortauthors{M. Ghader et~al.}  

\title [mode = title]{Forward-Only Convolutional Neural Networks with Learnable Channel–Class Assignment}  



%------------------------------------------------------------

% Author 1
\author[1]{Mohammadnavid Ghader}[%type=editor,
                        % auid=000,bioid=1,
                        % prefix=Sir,
                        % role=Researcher,
                        orcid=0009-0001-4465-4503]

% Email id
\ead{m_ghader@sbu.ac.ir}

\credit{Conceptualization, Methodology, Investigation, Formal analysis, Visualization, Writing – original draft.}

% Author 2 (Corresponding Author)
\author[1]{Saeed Reza Kheradpisheh}[%type=editor,
                        % auid=000,bioid=1,
                        % prefix=Sir,
                        % role=Researcher,
                        orcid=0000-0001-6168-4379]

\cormark[1]

% Email id
\ead{s_kheradpisheh@sbu.ac.ir}

\credit{Conceptualization, Methodology, Supervision, Writing – review and editing.}

% Author 3
\author[2]{Bahar Farahani}[%type=editor,
                        % auid=000,bioid=1,
                        % prefix=Sir,
                        % role=Researcher,
                        orcid=0000-0002-7016-6853]

% Email id
\ead{b_farahani@sbu.ac.ir}

\credit{Validation, Investigation, Writing – review and editing.}

% Author 4
\author[3]{Mahmood Fazlali}[%type=editor,
                        % auid=000,bioid=1,
                        % prefix=Sir,
                        % role=Researcher,
                        orcid=0000-0002-1701-5562]

% Email id
\ead{m.fazlali@herts.ac.uk}

\credit{Supervision, Funding acquisition, Project administration, Resources, Writing – review and editing.}

% Affiliations
\affiliation[1]{
    organization={Department of Computer and Data Science, Faculty of Mathematical Sciences, Shahid Beheshti University},
    city={Tehran},
    country={Iran}
}

\affiliation[2]{
    organization={Cyberspace Research Institute, Shahid Beheshti University},
    city={Tehran},
    country={Iran}
}

\affiliation[3]{
    organization={Cybersecurity and Computing Systems Research Group, University of Hertfordshire},
    city={Hatfield},
    country={United Kingdom}
}

\cortext[1]{Corresponding author}

\begin{abstract}
The Forward-Forward (FF) algorithm offers a biologically inspired alternative to backpropagation by replacing gradient-based credit assignment with local, forward-only objectives. While recent extensions have adapted FF to convolutional neural networks (CNNs), existing formulations rely on static channel–class partitions and struggle to perform effectively in complex tasks. In this work, we introduce a learnable channel–class assignment mechanism that enables adaptive, data-driven specialization of convolutional channels, supported by entropy and orthogonality regularization to promote learning performance. We further propose a loss-aware layer contribution strategy that adaptively weights intermediate-layer predictions based on their validation performance, enhancing the effectiveness of forward-only inference. Integrated into residual CNNs, the proposed method achieves consistently superior performance across CIFAR-10, CIFAR-100, and Tiny-ImageNet compared to existing similar forward-only methods. Notably, it establishes new state-of-the-art performance among FF-based models, substantially narrowing the gap with backpropagation. These findings demonstrate that introducing learnable channel specialization and layer contribution weighting significantly enhances the representational capacity of forward-only learning in deep CNNs.
\end{abstract}

%-------------------------------------------

% % Research highlights
% \begin{highlights}
% \item 
% \item 
% \item 
% \end{highlights}

%------------------------------------------


\maketitle

% Main text
%-----------------------------------------------------------
\section{Introduction}\label{sec1}

Deep learning has achieved remarkable success across diverse domains, including computer vision, natural language processing, and many other practical tasks, largely due to the effectiveness of gradient-based optimization via backpropagation \cite{rumelhart1986learning,krizhevsky2012imagenet,vaswani2017attention}. 
Despite its empirical achievements, backpropagation is constrained by several well-documented limitations. These include vanishing and exploding gradients in very deep networks \cite{bengio1994learning,glorot2010understanding}, dependence on full knowledge of forward computations that hinders black-box handling within the neural network, high memory requirements for storing activations during the forward and backward passes, and update locking due to the sequential nature of these passes \cite{huo2018decoupled}. Moreover, backpropagation faces several well-known challenges regarding biological plausibility \cite{lillicrap2016random}, such as reliance on global error signals, the assumption of symmetric weight transport between forward and backward paths, the requirement to freeze intermediate activity states until gradients are computed, and the separation of inference and learning.
These limitations have motivated the exploration of some alternatives that relax the strict requirements of global error transport while retaining strong representational power.

A growing body of research focuses on approaches that mitigate some of the challenges associated with backpropagation. Feedback alignment (FA) replaces precise feedback with random fixed matrices \cite{lillicrap2016random}, and direct feedback alignment (DFA) extends this to layer-wise updates \cite{nokland2016direct}. Variants such as weight mirror models \cite{akrout2019weight} refine the feedback to improve learning. Target propagation \cite{bengio2014auto,bartunov2018assessing} and its difference-based extensions \cite{lee2015difference,ernoult2022towards} assign local targets, while local representation alignment \cite{reclra} employs Hebbian-like top–down signals. Other strategies, including decoupled greedy learning \cite{belilovsky2020decoupled}, decoupled neural interfaces \cite{jaderberg2017decoupled,czarnecki2017understanding}, and PEPITA \cite{pepita}, introduce auxiliary networks, synthetic gradients, or perturbed forward passes to enable local updates. Although these methods relax the constraints of strict backpropagation, they often still rely on global signals and face challenges in scalability and accuracy.  

The recently proposed Forward–Forward (FF) algorithm introduces a paradigm that eliminates the backward pass entirely \cite{hinton2022forward}. Instead of propagating gradients, the original FF employs two forward passes. The first forward pass is conducted using real data, whereas the second employs corrupted or negative data, with the objective of optimizing a local goodness measure at each layer. By reframing training as a layer-wise contrastive process, FF achieves biologically inspired credit assignment without explicit global error backpropagation. However, the original formulation exhibits several practical limitations, including sensitivity to the choice of goodness measure and loss function, inefficiencies in multi-class classification, and difficulties in scaling to deeper architectures \cite{trifecta}. 

One potential limitation of the algorithm was that it operated on a fully connected architecture, which in turn imposed numerous constraints\cite{ghaderflff, ghadersnn}. Accordingly, several approaches were introduced that implemented the FF on convolutional neural networks. Recently, one of them was proposed by Sun et al~\cite{deeperff} that has sought to address the problem of network deepening in FF-based architectures and achieves state-of-the-art accuracy on several benchmarks using FF-based training models. This paper presents an approach to enhance the proposed FF-based CNN method and overcome some limitations of that. Our main contributions are the following:

\begin{itemize}
    \item We enhance the learning capacity of convolutional layers by allowing the contribution of channels to the classification process to be adaptively adjusted, rather than being statically assigned to specific classes.
    \item We propose a new approach for quantifying the contribution of individual layers to the overall model prediction, grounded in an evaluation of their localized performance.
    \item Proposed method demonstrates that learnable channel-class assignments enable more efficient channel utilization than static channel grouping, thereby achieving superior performance with significantly fewer channels.
\end{itemize}

The rest of this paper is organized as follows: Section~\ref{sec2} overviews related work and backgrounds of the FF algorithm. Section~\ref{sec3}  presents the proposed method. Section~\ref{sec4}  elaborates on the experimental setup and results. Finally, Section~\ref{sec5} concludes the paper and suggests a direction for future work. 

%%%%%%%%%%%%%%%%%%%%%%%%%%%%%%%%%%%%%%%%%%%%%%%%%%%%%

\section{Related Work and Background}\label{sec2}
This section provides an overview of the prior research and theoretical foundations that underpin our approach. We first review the fundamental principles of the Forward–Forward algorithm and its key developments, followed by recent efforts to extend this framework to convolutional architectures. Finally, we provide an overview of channel-wise convolution and explain its role in classification under forward-only training schemes.

\subsection{Forward-Forward Algorithm}
The Forward–Forward algorithm was proposed as a simple and biologically inspired alternative to backpropagation \cite{hinton2022forward}. Inspired by Boltzmann machines \cite{hinton1986boltzmann} and noise contrastive estimation \cite{gutmann2010noise}, FF introduces a greedy learning scheme where each layer maximizes the goodness of positive data, corresponding to correct or meaningful input associations, and minimizes the goodness of negative data, formed by incorrect or corrupted associations of the same inputs. This eliminates the backward pass and instead relies solely on local objectives. The development of the Forward–Forward algorithm has stimulated a diverse body of research, which can be organized around several key innovations and conceptual directions.

Some research has been conducted to improve and mitigate some challenges of the original Forward-Forward algorithm. 
SymBa~\cite{symba} addresses the problem of the asymmetric nature of the original FF algorithm by introducing a new loss function, which balances the positive and negative losses. Unlike the original FF algorithm, where the gradients for positive and negative samples behave differently, SymBa's loss function directly uses the difference in the goodness of positive and negative samples, making it symmetric. 
The Predictive Forward-Forward algorithm~\cite{PFF} introduces a credit assignment approach for neural systems by combining elements of predictive coding with the Forward-Forward algorithm. It features a recurrent neural network with a representation circuit for acquiring data representations and a generative circuit for synthesizing data.
One of the key challenges with the FF algorithm was its difficulty in generating compelling negative examples. In SCFF~\cite{SCFF}, positive examples are generated by concatenating identical data samples, while negative examples are constructed by concatenating samples from different classes.

Attention has also been devoted to biological plausibility and theoretical principles of the forward-forward algorithm. Terres-Escudero et al. \cite{neohebbian} linked FF to neo-Hebbian dynamics, reinforcing its alignment with theories of local synaptic plasticity and offering a biologically grounded interpretation of forward-only learning. Yang \cite{yang2023} provided a theoretical analysis of sparsity in FF activations, showing that the emergent representations resemble cortical coding patterns. 
Also, some studies~\cite{ghadersnn, terressnn, ororbiasnn} extended FF to event-driven spiking neural systems, confirming its potential for compatibility with temporal data and neuromorphic computation. 

\subsection{Convolutional Forward-Forward} \label{CFF}
Recent works have extended the Forward-Forward algorithm beyond fully connected networks by adapting it to convolutional architectures, addressing key challenges related to depth, feature extraction, and representational capacity. These efforts reinterpret the notion of goodness, introduce alternative local objectives, and modify training dynamics to better exploit spatial structure while preserving the forward-only, layerwise learning principle. The following approaches exemplify how convolutional designs significantly improve the performance of forward-only methods.

Dooms et al.~\cite{trifecta} introduced three synergistic techniques that address the limitations of the original Forward-Forward framework in scaling to deeper architectures. They replaced the threshold-dependent loss with the symmetric contrastive SymBa loss to improve stability, substituted length-based normalization with Batch Normalization to preserve inter-layer information and enhance generalization, and proposed Overlapping Local Updates to provide semi-local error propagation while maintaining layerwise independence. Collectively, these modifications significantly improved the depth scalability, accuracy, and stability of Forward-Forward training.

Zhao et al.~\cite{cafo} presented Cascaded Forward (CaFo), a biologically motivated alternative to backpropagation that employs a block-wise structure with locally attached predictors trained independently using objectives such as mean squared error, cross-entropy, or sparsemax. Instead of symmetric gradient transport, CaFo employs fixed random feedback for parameter updates, aligning with the biological plausibility of direct feedback mechanisms. Predictions from all predictors are aggregated at inference to form the final decision, thereby stabilizing classification and enhancing robustness. By eliminating the need for synthetic negative samples and surrogate goodness measures, CaFo reduces computational burden and introduces a probabilistic classification framework suitable for parallelized training.

Papachristodoulou et al.~\cite{cwconv} proposed CwComp, which reformulated the concept of goodness in convolutional layers by partitioning channels into class-specific subsets and comparing intra-class and inter-class activations. They introduced two losses: PvN loss, extending the original Forward-Forward principle, and Channel-wise Competition (CwC) loss, which uses softmax-based cross-entropy for layer-level classification. To operationalize this, they proposed the Channel-wise Feature Separator and Extractor (CFSE) block, combining standard and grouped convolutions to strengthen intra-class extraction and inter-class separation. Furthermore, they designed multiple prediction strategies and introduced Interleaved Layer Training to accelerate convergence and reduce stagnation.

Sun et al.~\cite{deeperff} identified key limitations in the CwComp~\cite{cwconv}, such as neuron deactivation and goodness leakage, which constrained its application to shallow models. To address these issues, they proposed DeeperForward, which replaces squared goodness with mean goodness and employs layer normalization to maintain consistent feature statistics. They extended the framework with channel-wise convolution structures, residual shortcuts, and the Signal Integrating module to selectively integrate predictions across layers. Training remains fully forward and local, with each layer independently optimized using cross-entropy loss. The approach supports model parallelism across GPUs and incorporates a memory-saving strategy by releasing intermediate states, thereby improving scalability and efficiency for deep architectures.

Despite the notable advancements introduced by CwComp~\cite{cwconv} and DeeperForward~\cite{deeperff}, both methods rely on predefined, static channel grouping mechanisms that permanently partition convolutional channels into fixed, class-specific subsets throughout training. In CwComp, grouped convolutions and channel-wise competitive learning explicitly enforce isolated class-dependent feature groups, while DeeperForward~\cite{deeperff} preserves this principle through channel-wise convolution (CW-Conv) structures that compute class-specific goodness scores from fixed channel partitions. Although these designs improve class separation and facilitate layer-wise local learning, they impose the restrictive assumption that semantic representations should remain permanently aligned with predetermined channel subsets. Such rigid assignments limit the network’s ability to dynamically adapt channel specialization to evolving feature distributions during training, leading to inefficient feature utilization and reduced representational flexibility. Moreover, fixed channel allocation implicitly assumes equal representational requirements across classes and restricts cross-group feature interaction, potentially hindering the learning of shared semantic structures and richer hierarchical representations in deeper architectures. Consequently, these static grouping strategies constrain the feature-learning procedure, motivating the development of a learnable channel-class assignment mechanism that enables channels to dynamically specialize based on data structure during optimization.

\subsection{Background of channel-wise convolution}
Traditional convolutional neural networks often learn features that overlap semantically across different channels, resulting in redundancy and weaker class discrimination. Channel-wise convolution (CW-Conv)~\cite{cwconv} mitigates this issue by explicitly partitioning the feature space, thereby enabling two key mechanisms: \textit{intra-class feature specialization} and \textit{inter-class competition}. Through intra-class specialization, each subset of channels focuses on learning features that are most relevant to a specific class. In contrast, inter-class competition encourages different channel groups to compete for activation, enhancing the model’s discriminative capacity. Let the feature map output of a convolutional layer be denoted as $A \in \mathbb{R}^{C \times H \times W}$, where $C$ is the number of channels, and $H$ and $W$ represent the spatial dimensions. For a dataset containing $J$ distinct classes, the channel dimension is divided into $J$ disjoint subsets, each corresponding to one class. Specifically, let $A_i \in \mathbb{R}^{S \times H \times W}$, where $S = C / J$, represent the subset of feature maps assigned to the $i$-th class.
The \textit{goodness} score of each subset is defined as the spatial and channel-wise mean of the squared activations:
\begin{equation}
G_{i} = \frac{1}{S \times H \times W} 
\sum_{s=1}^{S} \sum_{h=1}^{H} \sum_{w=1}^{W} 
A_{i}^{2},
\end{equation}
where $G_{i}$ quantifies the activation strength of the $i$-th class subset. Collectively, these values form $G \in \mathbb{R}^{J}$, which represents the per-class activation strengths.

%%%%%%%%%%%%%%%%%%
Because activations are passed through ReLU, some neurons output zero whenever their pre-activation is negative. When squared goodness is used, the gradient formulation for each neuron becomes proportional to that neuron’s activation value. This implies that when a neuron produces zero output, its corresponding gradient is likewise zero. As a result, the incoming weights of that neuron receive no update, so the neuron remains inactive in learning procedure, leading to what is called the deactivated neuron problem.

\begin{figure}
\centering
\includegraphics[width=0.99\textwidth]{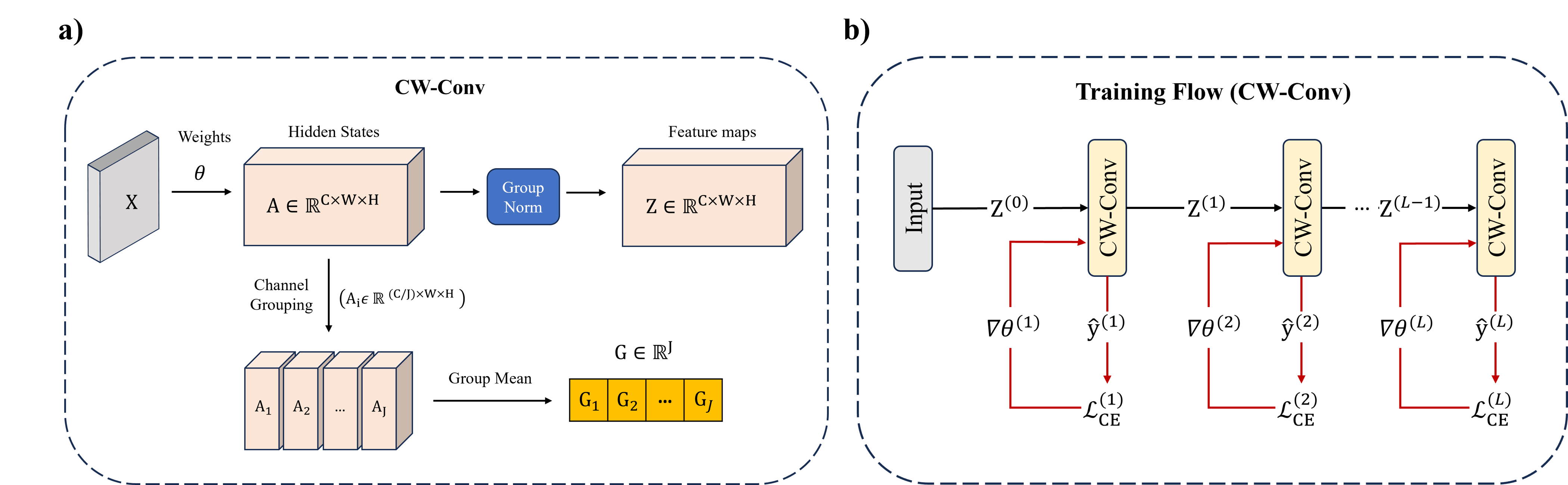}
% \vspace{1em}
\caption{Overview of layers and training scheme of DeeperForward~\cite{deeperff}. (a) Structure of the CW-Conv block, illustrating grouped channel processing. (b) Training flow of stacked CW-Conv layers, and the local weight updating using layer-wise loss.}

\label{fig:CWConv}
\end{figure}

% In another study~\cite{deeperff}, several modifications are introduced. As shown in Fig.~\ref{fig:CWConv}(a), for the output of a convolutional layer after ReLU activation $A \in \mathbb{R}^{C \times H \times W}$, the feature channels are evenly divided into \(J\) groups, each associated with one of the \(J\) target classes. Thus, each subset of feature maps \(A_i \in \mathbb{R}^{S \times H \times W}\) corresponds to class \(i\). For calculating the goodness of each class, mean goodness is utilized.
In another study~\cite{deeperff}, several modifications are introduced. As shown in Fig.~\ref{fig:CWConv}(a), for the output of a convolutional layer after ReLU activation $A$, the feature channels are evenly divided into \(J\) groups, each associated with one of the \(J\) target classes. Thus, each subset of feature maps \(A_i \in \mathbb{R}^{S \times H \times W}\) corresponds to class \(i\). For calculating the goodness of each class, mean goodness is utilized.
For each group \(i\), the goodness is calculated as the average magnitude of activation across the spatial and channel dimensions:

\begin{equation}
\label{eq:goodness}
{G}_i = \frac{1}{S \times H \times W} 
\sum_{s=1}^{S} \sum_{h=1}^{H} \sum_{w=1}^{W} A_{i}.
\end{equation}

The class-wise goodness is represented as ${G} = [G_1, G_2, \dots, G_J] \in \mathbb{R}^J$, which is used to perform layer-wise classification. 
% Using the mean of activations in equation~\ref{eq:goodness} ensures that all neurons contribute to each update of layer parameters, thereby mitigating the deactivated neuron problem.
After computing \(G\), the softmax function is applied to obtain class probabilities for prediction and loss computation:
\begin{equation}
\hat{y} = \mathrm{softmax}(G).
\end{equation}

As shown in Fig.~\ref{fig:CWConv}(b), the training procedure in the mentioned framework follows the principles of the forward-only algorithms, thereby eliminating the need for a backward error-propagation phase. Each CW-Conv layer is optimized independently using a local objective that depends solely on its input and output representations. The cross-entropy (CE) loss employed for local classification in both~\cite{cwconv,deeperff} is defined as:
\begin{equation}
\label{eq:CE}
\mathcal{L}_{CE}(\hat{y}, y)
= -\sum_{i=1}^{J} y_i \log(\hat{y}_i),
\end{equation}
where \(y\) represents the corresponding one-hot encoded ground-truth label.
Before propagating the features to the subsequent layer, it is necessary to decouple their scale from the representational direction. Therefore, the feature maps are normalized to remove class-specific activation magnitude. The normalized representation is defined as $Z = \text{GroupNorm}(A; J)$, where \(\text{GroupNorm}(\cdot)\) performs normalization within each channel group.

As discussed in the~\ref{CFF}, the static assignment of channel groups to different classes, along with their fixed associations throughout the training process, may lead to underutilization of network capacity. By introducing a learnable component that determines each channel's contribution to decision-making across classes at each layer, the predefined channel grouping mechanism can be replaced with an adaptive process continuously optimized during training.

% \vspace{1em}
\noindent

%%%%%%%%%%%%%%%%%%%%%%%%%%%%%%%%%%%%%%%%%%%%%%%%%%%%%
\section{Proposed Method}\label{sec3}
This section describes our approach to train convolutional neural networks with a forward-only algorithm. The mathematical formulations used at each step of the proposed method are also presented.

\subsection{Convolutional Layers}
\begin{figure}
\centering
\includegraphics[width=0.95\textwidth]{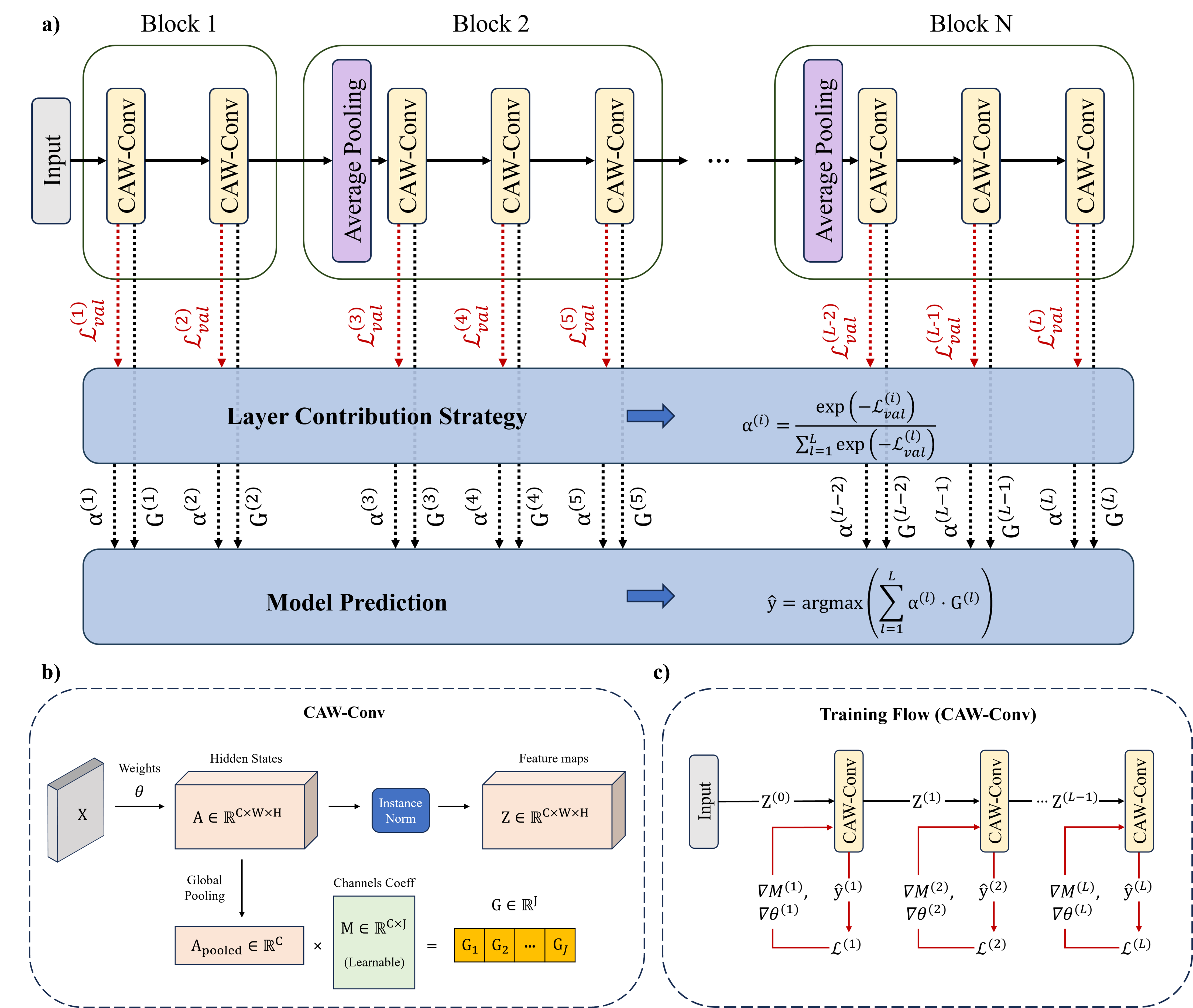}
% \vspace{1em}
% \caption{Overview of the proposed forward-only method for training of CNNs, illustrating the overall architecture, adaptive layer contribution and prediction strategy, and the internal design of the FF-Conv block.}
\caption{Overview of the proposed forward-only learning framework for CNNs. (a) Overall architecture of the model, showing stacked CAW-Conv blocks, and the layer contribution strategy used for final decision making. (b) Internal structure of the CAW-Conv block. (c) Training flow of the network, illustrating forward propagation and layer-wise cross-entropy loss optimization.}
\label{fig:Main}
\end{figure}

The channel-wise idea of convolutional layers proposed by~\cite{cwconv}, in which the convolutional layers are designed to assign specific groups of channels to each class, encouraging them to be activated when encountering corresponding class samples. The channels are statically preserved within their initially assigned groups during training, contributing exclusively to a specific class. In this work, we seek to remove this static class-channel assignment and propose a dynamic assignment strategy instead.
As shown in Fig.~\ref{fig:Main}(b), \( x \in \mathbb{R}^{C_{\text{in}} \times H \times W} \) denote the input tensor to the convolutional layer, where \( C_{\text{in}} \) is the number of input channels, and \( H \times W \) represents the spatial dimensions. The layer produces hidden states \( A \in \mathbb{R}^{C_{\text{out}} \times H' \times W'} \), where \( C_{\text{out}} \) is the number of output channels, and \( H' \times W' \) are the output spatial dimensions determined by the convolution parameters such as kernel size, stride, and padding.
The convolutional layer applies a standard 2D convolution followed by a ReLU activation to construct hidden states $A$:
\begin{equation}
% A = \text{Dropout}(\text{ReLU}(\text{Conv2D}(x; W, b))),
A = \text{ReLU}(\text{Conv2D}(x; \theta)),
\end{equation}
where \( \theta \in \mathbb{R}^{C_{\text{out}} \times C_{\text{in}} \times K \times K} \) is the convolutional weight tensor, \( b \) is the bias, and \( K \) is the kernel size.
To enable adaptive class-channel assignment, we introduce a learnable parameter matrix \( U \in \mathbb{R}^{C_{\text{out}} \times J} \), which is randomly initialized, where \( J \) denotes the number of classes. Each row of the matrix \( U \) denotes the contribution rate of each channel to the channel-wise convolution for each class. To set a threshold for the contribution of each channel across different classes, the matrix \( M \) is constructed by applying the softmax function to the row elements of \( U \):
%with a temperature parameter \( T \):
\begin{equation}
% M^{(soft)}_{i,j} = \frac{\exp(M_{i,j})}{\sum_{j=1}^J \exp(M_{i,j})}.
M_{i,j} = \frac{\exp(U_{i,j})}{\sum_{j=1}^J \exp(U_{i,j})}.
% M^{'}_{i,j} = \frac{\exp(M_{i,j})}{\sum_{j=1}^J \exp(M_{i,j})}.
\end{equation}
The softmax ensures that the weights for each channel sum to 1 across all classes, i.e., \( \sum_{j=1}^J M_{i,j} = 1 \). 
Since each layer can perform its own independent decision-making process, this decision is evaluated based on the activation of neurons across different classes. Accordingly, the goodness \( G \in \mathbb{R}^{J} \), representing the class-specific output, is defined as follows:
\begin{equation}
G = A_{\text{pooled}} \cdot M,
\end{equation}
where \( A_{\text{pooled}} \in \mathbb{R}^{C_{\text{out}}} \) is a pooled representation of the current layer feature maps, defined as a weighted combination of mean, maximum, and minimum neurons' activation over the spatial dimensions:
\begin{equation}
% A_{\text{pooled}} = \beta \cdot (\sum_{h=1}^{H} \sum_{w=1}^{W} {A}_{h,w}) + (1-\beta) \cdot (\max_{h,w} {A}_{h,w} - \min_{h,w} {A}_{h,w})
% A_{\text{pooled}} = \beta \cdot (\frac{1}{H\times W}\sum_{h=1}^{H} \sum_{w=1}^{W} {A}) + (1-\beta) \cdot (\max_{h,w} {A} - \min_{h,w} {A}),
% A_{\text{pooled}} = \beta \cdot (\mean_{h,w} {A}) + (1-\beta) \cdot (\max_{h,w} {A} - \min_{h,w} {A}),
% A_{\text{pooled}} = \beta \cdot \left( \operatorname*{mean}_{h,w} A \right) + (1-\beta) \cdot \left( \operatorname*{max}_{h,w} A - \operatorname*{min}_{h,w} A \right)
A_{\text{pooled}} = \beta \cdot \operatorname*{mean}_{h,w} A + (1-\beta) \cdot ( \operatorname*{max}_{h,w} A - \operatorname*{min}_{h,w} A)
\end{equation}
where $\beta \in [0,1]$ is a scalar weighting coefficient that controls the trade-off between the average activation component and the extremal activation component. This pooling strategy captures both average and extreme feature activations to enhance the extracted information from channel values.
The local prediction of each layer is computed based on $G$ the following:
\begin{equation}
\hat{y} = \mathrm{softmax}(G).
\end{equation}

As representation from each CW-Conv module serves as the input for the next layer, for decoupling scale from feature maps $A$, the instance normalization~\cite{IN} is applied as follows:

\begin{equation}
% Z_{c,h,w} = \gamma_c \cdot \frac{A_{c,h,w} - \mu_{c}}{\sqrt{\sigma_{c}^2 + \epsilon}} + \beta_c,
% Z = \gamma \cdot \frac{A - \mu_{c}}{\sqrt{\sigma_{c}^2 + \epsilon}} + \beta,
Z_{c,h,w} = \gamma_c , \frac{A_{c,h,w} - \mu_c}{\sqrt{\sigma_c^2 + \epsilon}} + \beta_c,
\label{eq:inorm}
\end{equation}
where the mean $\mu_{c}$ and variance $\sigma_{c}$ are computed over the spatial dimensions \( H \) and \( W \):

\begin{equation}
\begin{aligned}
% \mu_{c} &= \frac{1}{H \times W} \sum_{h=1}^{H} \sum_{w=1}^{W} A_{c}, \\
\mu_c &= \frac{1}{H W} \sum_{h=1}^{H} \sum_{w=1}^{W} A_{c,h,w},, \\
% \sigma_{c}^2 &= \frac{1}{H \times W} \sum_{h=1}^{H} \sum_{w=1}^{W} (A_{c} - \mu_{c})^2,
\sigma_c^2 &= \frac{1}{H W} \sum_{h=1}^{H} \sum_{w=1}^{W} \left( A_{c,h,w} - \mu_c \right)^2,
\end{aligned}
\label{eq:stats}
\end{equation}
where \( \epsilon \) is a small constant for avoiding zero division, and \( \gamma_c \) and \( \beta_c \) are learnable parameters corresponding to scaling and shifting for each channel \( c \). Therefore, the normalized feature maps $Z$ are then supplied as input to the subsequent layer.

\subsection{Loss calculation}
After completing the forward-pass computations and computing the prediction vector $\hat{y}$ for a given layer, the corresponding loss terms can be evaluated. The primary objective is the cross-entropy loss $L_{CE}$, which quantifies the discrepancy between predicted class probabilities and ground-truth labels as described in equation~(\ref{eq:CE}). To further encourage channel-level specialization such that different channels capture discriminative features for distinct classes, we incorporate two additional regularization terms into the loss function:
\begin{enumerate}
    \item \textbf{Entropy Regularization}: We minimize the entropy of the class-channel weights to encourage sparsity, ensuring that each channel contributes significantly to only a few classes:
    \begin{equation}
    \mathcal{L}_{\text{ent}} = -\frac{1}{J} \sum_{j=1}^J \sum_{i=1}^{C_{\text{out}}} M_{i,j} \log(M_{i,j} + \epsilon),
    \end{equation}
    where \( \epsilon \) is a small value that prevents numerical instability.
    \item \textbf{Orthogonality Regularization}: To promote diversity among class-specific channel assignments, we enforce orthogonality on the normalized weights. Let \( M_{\text{norm}} = M / \|M\|_2 \) be the column-normalized matrix. The Gram matrix is computed as \( Gram = M_{\text{norm}}^T M_{\text{norm}} \), and we minimize the deviation from the identity matrix:
    \begin{equation}
    \mathcal{L}_{\text{ortho}} =  \| Gram - I_J \|_F^2,
    %\frac{1}{M^2 C_{\text{out}}}
    \end{equation}
    %\cdot \min\left(1, \frac{\text{epoch}}{50}\right)
    where \( I_J \) is the \( J \times J \) identity matrix, and the Frobenius norm \( \| \cdot \|_F \) measures the deviation. 
    %The curriculum factor \( \min(1, \text{epoch}/50) \) gradually increases the regularization strength over the first 50 epochs.
\end{enumerate}
The total loss for the CAW-Conv layer is:
\begin{equation}
\mathcal{L} = \mathcal{L}_{\text{CE}} + \lambda_{\text{ent}}\mathcal{L}_{\text{ent}} + \lambda_{\text{ortho}}\mathcal{L}_{\text{ortho}},
\end{equation}
where $\lambda_{ent}$ and $\lambda_{ortho}$ denote the weighting coefficients that control the strength and the contributions of the entropy and orthogonality regularization terms to the layer loss, respectively.
%%%%%%%%%%%%%%%%%%%%%%%%%%%%%%%%%%%%%%%%%%%%%%%%%%%%%%%%%%%%%
Although the training process in different layers is conducted independently, deeper layers utilize the features extracted by preceding layers in order to derive higher-level representations. Fig.~\ref{fig:Main}(c), illustrates the training flow of the multi-layer CAW-Conv network. 
Given an input, an initial representation \( Z^{(0)} \), corresponding to a normalized version of the input, is first obtained and subsequently processed through \( L \) stacked CAW-Conv layers.
At layer $\ell$, the previous layer output $Z^{(\ell-1)}$ is transformed into a new feature representation $Z^{(\ell)}$ using the convolutional weights $\theta^{(\ell)}$ and the channel-class matrix $M^{(\ell)}$.

% Unlike conventional architectures that only produce a prediction at the final layer, each CAW-Conv layer generates its own prediction $\hat{y}^{(\ell)}$. Each layer prediction is supervised independently by computing a layer-specific loss $\mathcal{L}^{(\ell)}$.

During the weight updating phase, each layer loss $\mathcal{L}^{(\ell)}$ produces gradients with respect to both convolutional weights ($\nabla_{\theta^{(\ell)}} \mathcal{L^{(\ell)}}$) and the channel-class matrix ($\nabla_{M^{(\ell)}} \mathcal{L^{(\ell)}}$), which are written in abbreviated form as $\nabla{\theta^{(\ell)}}$ and $\nabla{M^{(\ell)}}$, respectively.
These gradients are used to update the learnable parameters of the corresponding layer. 
The illustrated training flow shows that the optimization does not rely solely on the final-layer loss, and every layer has its own local weight updating based on its internal error.

%%%%%%%%%%%%%%%%%%%%%%%%%%%%%%%%%%%%%%%%%%%%%%%%%%%%%%%%%%%%%

%%%%%%%%%%%%%%%%%%%%%%%%%%%%%%%%%%%%%%%%%%%%%%
\begin{algorithm}[h]
\caption{Single Forward Pass training with Layer-wise Class-Adaptive Weighted Convolution and Inference}
\label{alg:cwc_resnet}
\begin{algorithmic}[1]
\Require Training set $\mathcal{D}_{\text{train}}$, validation set $\mathcal{D}_{\text{val}}$, test set $\mathcal{D}_{\text{test}}$, number of layers $L$, number of epochs $E$
\Function{CAW-Conv}{$x, \theta, M$}
    \State $A = \text{ReLU}(\text{Conv2D}(z^{(l-1)}; \theta))$ 
    \State $A_{\text{pooled}} = \beta \cdot \operatorname*{mean}_{h,w} A + (1-\beta) \cdot \left( \max_{h,w} A - \min_{h,w} A \right)$
    \State $G \gets A_{\text{pooled}} \cdot M$ 
    \State $Z \gets \text{InstanceNorm}(A)$ 
    \State \textbf{return} $Z,\ G$
\EndFunction
% \State \Call{Partition1}{}
\Procedure{Training}{}
    \For{each layer $l=1$ to $L$}
        \State Initialize $U^{(l)}$, $\theta^{(l)}$
    \EndFor
    \For{each epoch $e = 1$ to $E$} 
        \For{each layer $l=1$ to $L$}
            % \State $M_{i,j}^{(l)} \gets \text{Softmax}(U_{i,j}^{(l)})$
            % \State $A = \text{ReLU}(\text{Conv2D}(z^{(l-1)}; \theta^{(l)}, b^{(l)}))$ 
            % \State $A_{\text{pooled}}^{(l)} = \beta \cdot \operatorname*{mean}_{h,w} A + (1-\beta) \cdot \left( \max_{h,w} A - \min_{h,w} A \right)$
            % \State $G^{(l)} \gets A_{\text{pooled}} \cdot M^{(l)}$ %\Comment{Class prediction logits from layer $i$}
            % \State $Z^{(l)} \gets \text{InstanceNorm}(A)$ %\Comment{Feature normalization}
            % \State \textbf{return} $Z,\ G^{(l)}$
            \State $M^{(l)} \gets \text{Softmax}(U^{(l)})$
            \State $Z^{(l)},\ G^{(l)} \gets CAW-Conv(Z^{(l-1)}, \theta^{(l)}, M^{(l)})$

            \State Entropy loss: $\mathcal{L}_{\text{ent}} = - \frac{1}{J} \sum_{j=1}^J \sum_{i=1}^{C_{\text{out}}} M_{i,j} \log(M_{i,j} + \epsilon)$
            \State Orthogonality loss: $\mathcal{L}_{\text{ortho}} = \|M_{\text{norm}}^\top M_{\text{norm}} - I_J\|_F^2$
            \State $\mathcal{L}^{(l)} \gets \mathcal{L}_{\text{CE}}^{(l)} + \lambda_{ent}^{(l)}\mathcal{L}_{\text{ent}}^{(l)} + \lambda_{ortho}^{(l)}\mathcal{L}_{\text{ortho}}^{(l)}$
            \State Update $U^{(l)}$ and $\theta^{(l)}$ based on $\mathcal{L}^{(l)}$   
        \EndFor
    \EndFor
\EndProcedure
\Procedure{Inference}{}
    \For{layer $l = 1$ to $L$} \hspace{0.5em} // Layer Contribution Strategy 
        \State \( \alpha^{(i)} = \frac{\exp(-{\mathcal{L}}_{val}^{(i)})}{\sum_{l=1}^L \exp(-{\mathcal{L}}_{val}^{(l)})} \). 
    \EndFor
    \State \( G_{\text{total}} = \sum_{l=1}^L \alpha^{(l)} \cdot \text{G}^{(l)} \) \hspace{0.5em} // Model Prediction
    \State \(\hat{y} = \arg\max(G_{\text{total}}) \) 
\EndProcedure
\end{algorithmic}
\end{algorithm}
%%%%%%%%%%%%%%%%%%%%%%%%%%%%%%%%%%%%%%%%%%%%%%%%%%%%%

% \subsection{Layer Contribution Strategy}
\subsection{Inference}
In contrast to conventional neural networks, where the decision-making process is performed exclusively at the final layer, the decision-making mechanism in the proposed architecture can be carried out at each individual layer or through the integration of information obtained from multiple layers. Since different layers possess varying levels of representation learning, they should not necessarily contribute equally to the final decision-making process. Hence, we propose a strategy that dynamically adjusts each layer's contribution based on its average loss. 
% The algorithm evaluates the performance of each layer and weights its contributions to the final classification output.
% \noindent

\vspace{0.5em}
\textbf{Layer Contribution Strategy.} As described in Algorithm~\ref{alg:cwc_resnet}, there is \textit{Layer Contribution Strategy}, which combines the outputs of multiple layers in a neural network using a loss-aware weighting mechanism. The central idea is to exploit intermediate representations from different layers while adaptively weighting them based on their predictive reliability, as estimated from the validation set.  

Let the model consist of \(L\) layers, and let \(\mathcal{L}_{val}^{(\text{i})}\) denote the set of loss values computed from the \(i\)-th layer over the validation set \(\mathcal{D}_{\text{val}}\), which serves as a proxy for the quality of the layer’s feature representation. A smaller \(\mathcal{L}_{val}^{(\text{i})}\) indicates better alignment with the ground-truth labels, and hence the layer should receive higher importance in the final prediction.  

To translate these loss values into normalized importance scores, a softmax transformation is applied:
\begin{equation}
    \alpha^{(i)} = \frac{\exp(-\mathcal{L}_{val}^{(\text{i})})}{\sum_{l=1}^L \exp(-\mathcal{L}_{val}^{(\text{i})})}, \quad i=1,\dots,L.
\end{equation}
Here, the negative sign ensures that layers with lower losses are assigned larger weights, while the denominator normalizes the weights so that \(\sum_{i=1}^L \alpha^{(i)} = 1\).

% \vspace{0.5em}
\textbf{Model Prediction.} During inference, for a given input \(x\), each layer produces a goodness \(\text{G}^{(i)}\). These outputs are aggregated using the learned weights:
\begin{equation}
    G^{\text{(Total)}} = \sum_{l=1}^L \alpha^{(l)} \cdot G^{(l)},
\end{equation}
where \(G^{\text{(Total)}}\) denotes the weighted combination of class scores across all layers. The final prediction is then obtained by selecting the class with the maximum aggregated score:
\begin{equation}
    \hat{y} = \arg\max \big(G^{\text{(Total)}}\big).
\end{equation}
% The layer contribution strategy provides an adaptive mechanism that dynamically favors layers with higher discriminative power, effectively mitigating the risk of relying solely on the final layer and thereby leveraging rich intermediate representations. 

In general, by introducing a learnable matrix \( U \) at each layer that determines the contribution of each channel to the extraction of class-specific features, the training process can achieve more effective utilization of channel capacity than static grouping-based approaches. Furthermore, the proposed layer contribution strategy enables different layers to contribute to the final model prediction based on their performance and discriminative power, potentially improving the model's overall predictive performance.

%%%%%%%%%%%%%%%%%%%%%%%%%%%%%%%%%%%%%%%%%%%%%%%%%%%%%
\section{Experimental Results}\label{sec4}
This section presents the experimental evaluation of the proposed method. We first describe the datasets and experimental settings used to assess the performance of the model. Then, we provide a comprehensive comparison with existing approaches to demonstrate the effectiveness and competitiveness of the proposed framework.

\subsection{Datasets and experiment settings}
This study evaluates model performance on five benchmark datasets: MNIST~\cite{mnist}, Fashion-MNIST~\cite{fmnist}, CIFAR-10~\cite{cifar}, CIFAR-100~\cite{cifar}, and Tiny-ImageNet~\cite{tiny}. For MNIST and Fashion-MNIST, the 60,000 training samples are partitioned into subsets of 50,000 and 10,000 samples, while CIFAR-10 and CIFAR-100 split their 50,000 training samples into subsets of 40,000 and 10,000 samples. Tiny-ImageNet presents a more challenging benchmark, comprising 100,000 training images and 10,000 validation images across 200 object categories. In all cases, the validation subsets are used for the layer contribution strategy.

\begin{table}
\centering
\caption{Comparison of classification accuracy (\%) of BP-based, non-FF, and FF-based methods on CIFAR-10, MNIST, and Fashion-MNIST. $\dagger$: indicates results obtained with data augmentation.}

\label{table:results1}
\begin{center}
% \resizebox{0.9\textwidth}{!}{
\begin{tabular}{ccccccc}
\hline
\textbf{Type} & \textbf{Method} & \textbf{Arch.} & \textbf{\#Layer} & \textbf{CIFAR-10} & \textbf{MNIST} & \textbf{F-MNIST} \\[1pt]
\hline
\multirow{4}{*}{Block-wise BP} \\[-9pt]
& HPFF\cite{hpff} & CNN & 110 & 91.04 & - & - \\
& SEDONA\cite{sedona} & CNN & 152 & 93.87 & - & - \\
& BWBPF\cite{bwbpf} & CNN & 152 & \textbf{95.52} & - & - \\
& DF-O\cite{DF} & CNN & 10 & 88.15 & \textbf{99.70} & \textbf{93.89} \\
\hline
\multirow{1}{*}{Conventional BP} \\[-9pt]
& BP\cite{deeperff} & CNN & 18 & 94.03$\dagger$ & {99.58} & 93.78 \\
\hline
\multirow{4}{*}{non-FF (BP-Free)}
& PEPITA\cite{pepita} & CNN & 2 & 52.57 & 98.01 & - \\
& DTP\cite{ernoult2022towards} & CNN & 6 & \textbf{89.38} & 98.93 & \textbf{90.35} \\
% & recLRA\cite{reclra} & CNN & 18 & 93.58 & 98.18 & 88.13 \\
& SoftHebb\cite{softhebb} & SoftHebb & 4 & 80.31 & \textbf{99.35} & - \\
& F$^3$\cite{f3} & MLP & 2 & 46.04 & 97.16 & - \\
% & SP\cite{sp} & CNN & 8 & 92.4 & - & - \\
\hline
\multirow{6}{*}{FF-based (Shallow)} 
& FF\cite{hinton2022forward} & MLP & 4 & 59.00 & 98.69 & - \\ 
& SymBa\cite{symba} & MLP & 3 & 59.09 & 98.58 & - \\
& CaFo\cite{cafo} & CNN & 3 & 67.43 & 98.80 & - \\
& CwComp\cite{cwconv} & CNN & 4 & 78.11 & 99.42 & 92.31 \\
& DeeperForward\cite{deeperff} & CNN & 4 & 79.49 & 99.50 & 91.83 \\
% & Our method & CNN & 4 & \textbf{84.59} & \textbf{99.67} & \textbf{93.38} \\
& \textbf{Proposed method} & CNN & 4 & \textbf{84.59} & \textbf{99.67} & \textbf{93.38} \\
\hline
\multirow{7}{*}{FF-based (Deep)}
& DF-R\cite{DF} & CNN & 10 & 84.75 & 99.53 & 92.5 \\
& Trifecta\cite{trifecta} & CNN & 12 & 83.51 & 99.58 & 91.44 \\
& CwComp\cite{cwconv} & CNN & 14 & 75.28 & 99.27 & 91.79 \\
& DeeperForward\cite{deeperff} & CNN & 14 & 81.76 & 99.65 & 92.44 \\
% & DeeperForward\cite{deeperff} & CNN (VGG) & 14 & 81.76 & 99.65 & 92.44 \\
% & \textbf{Proposed method} & CNN (VGG) & 14 & 86.91 & 99.72 & 93.99 \\
& \textbf{Proposed method} & CNN & 14 & 86.91 & 99.72 & 93.99 \\
% & PLFF\cite{plff} & CNN & 17 & 87.02 & 99.80 & 93.68 \\
& DeeperForward\cite{deeperff} & CNN & 17 & 86.22 & 99.63 & 93.13 \\
% & DeeperForward\cite{deeperff} & CNN (ResNet) & 17 & 86.22 & 99.63 & 93.13 \\
& \textbf{Proposed method} & CNN & 17 & \textbf{89.37} & \textbf{99.74} & \textbf{94.55} \\
% & \textbf{Proposed method} & CNN (ResNet) & 17 & \textbf{89.37} & \textbf{99.73} & \textbf{94.55} \\
\hline
\end{tabular}
\end{center}
\end{table}
\begin{table}
\centering
\caption{Comparison of classification accuracies (\%) achieved by BP-based and forward-only methods on CIFAR-100 and Tiny-ImageNet. (CH$\times3$ indicates a model with three times the channel size.) \mbox{*} Reproduced results.}

\label{table:results2}
\begin{tabular}{lcc}
\hline
 & \textbf{CIFAR-100} & \textbf{Tiny-ImageNet} \\
\hline
BP\cite{cifar100ref, tinyref} & 70.70 & 61.8 \\
DeeperForward\cite{deeperff} & 53.09 & 41.36\mbox{*} \\
DeeperForward-CHx3\cite{deeperff} & 60.28 & - \\
\textbf{Proposed method} & 63.52 & 49.87 \\
\textbf{Proposed method-CHx3} & 69.74 & - \\
\hline
\end{tabular}
\end{table}

The proposed model is built upon three distinct architectures: TinyCNN-4, VGG-14, and ResNet-17, each designed with specific characteristics for feature extraction and representation learning.
TinyCNN is a compact convolutional network consisting of four CAW-Conv layers with progressively increasing channel sizes. The network starts with 100 channels in the first CAW-Conv layer, followed by a second CAW-Conv layer with 200 channels. The third CAW-Conv layer expands to 400 channels, and the final CAW-Conv layer retains 400 channels, ensuring a balance between efficiency and representational capacity.
VGG follows a traditional VGG-style architecture with 14 CAW-Conv layers organized into five blocks. The channel sizes increase progressively across the blocks: the first block contains 70 channels, the second includes 140, the third contains 280, the fourth contains 560, and the final block also maintains 560 channels, enabling deep feature extraction while maintaining manageable complexity.
ResNet consists of 17 CAW-Conv layers arranged into four blocks, with channel sizes increasing as follows: 100 in the first block, 200 in the second, 400 in the third, and 800 in the fourth. The architecture leverages residual connections, including additive and concatenation shortcuts, and uses average pooling for downsampling when needed, facilitating efficient learning even at deeper layers.
Each of these architectures uses a consistent optimization strategy, with all layers independently optimized using the AdamW optimizer and a cosine annealing learning rate schedule to enhance performance and convergence.
Additionally, the values of $\lambda_{ent}$ and $\lambda_{\text{ortho}}$ are kept fixed across all layers and set to $10^{-2}$ and $10^{-3}$, respectively.

\subsection{Comparisons of Different Methods}

%***********************
Table~\ref{table:results1} compares the proposed method with a diverse set of learning paradigms, including block-wise backpropagation methods, fully backpropagation-free approaches, and existing FF-based models.
Among shallow FF-based models, the proposed method with a 4-layer architecture (TinyCNN-4) achieves the highest accuracy on all three datasets, reaching 84.59\% on CIFAR-10, 99.67\% on MNIST, and 93.38\% on Fashion-MNIST. Compared with the strongest previous shallow FF baseline, DeeperForward~\cite{deeperff}, the proposed method improves CIFAR-10 accuracy by more than 5\%. It also substantially outperforms the original FF algorithm~\cite{hinton2022forward}, SymBa~\cite{symba}, CaFo~\cite{cafo}, and CwComp~\cite{cwconv}.

The superiority of the proposed approach is still evident as network depth increases. In the 14-layer architecture (VGG-14), the proposed model attains 86.91\%, 99.72\%, and 93.99\% accuracy on CIFAR-10, MNIST, and Fashion-MNIST, respectively, outperforming the corresponding DeeperForward~\cite{deeperff}. Similar trends are observed in the 17-layer architecture (ResNet-17), where the proposed method reaches 89.37\%, 99.74\%, and 94.55\%. These results indicate that the proposed learnable channel-class assignment and layer contribution mechanisms effectively exploit the additional representational capacity provided by deeper architectures, whereas many of existing FF methods benefit less from increased depth.

Compared with other deep FF-based approaches, including DF-R~\cite{DF}, Trifecta~\cite{trifecta}, and CwComp~\cite{cwconv}, the proposed model consistently achieves the highest accuracy across all evaluated datasets. In particular, the proposed 17-layer network surpasses DF-R~\cite{DF} by 4.62 percentage points on CIFAR-10 and 2.05 percentage points on Fashion-MNIST, while also improving MNIST accuracy. Furthermore, it outperforms Trifecta~\cite{trifecta} by 5.86, 0.16, and 3.11 percentage points on CIFAR-10, MNIST, and Fashion-MNIST, respectively. These findings demonstrate the superior scalability of the proposed framework and its ability to learn more discriminative representations under the FF learning paradigm.

When compared with non-FF backpropagation-free approaches, the proposed method remains highly competitive. On CIFAR-10, the proposed ResNet-17 model achieves an accuracy of 89.37\%, essentially matching the performance of DTP~\cite{ernoult2022towards} (89.38\%), which represents the strongest non-FF backpropagation-free baseline in the table. Moreover, on MNIST and Fashion-MNIST, the proposed method establishes new best results among all reported backpropagation-free approaches. Among BP-based methods, block-wise backpropagation approaches generally achieve higher CIFAR-10 accuracies, but they rely on partial error backpropagation and often require substantially deeper architectures. For example, BWBPF~\cite{bwbpf} attains 95.52\% accuracy using a 152-layer network, whereas the proposed method achieves 89.37\% using only 17 layers and without any backward error propagation. Notably, on MNIST and Fashion-MNIST, the proposed model surpasses both conventional backpropagation (BP)~\cite{deeperff} and DF-O~\cite{DF}. These results not only close the gap with BP but also demonstrate the competitiveness of FF-based learning on standard vision benchmarks. 
%***********************
\begin{figure}
\centering
\includegraphics[width=0.98\textwidth]{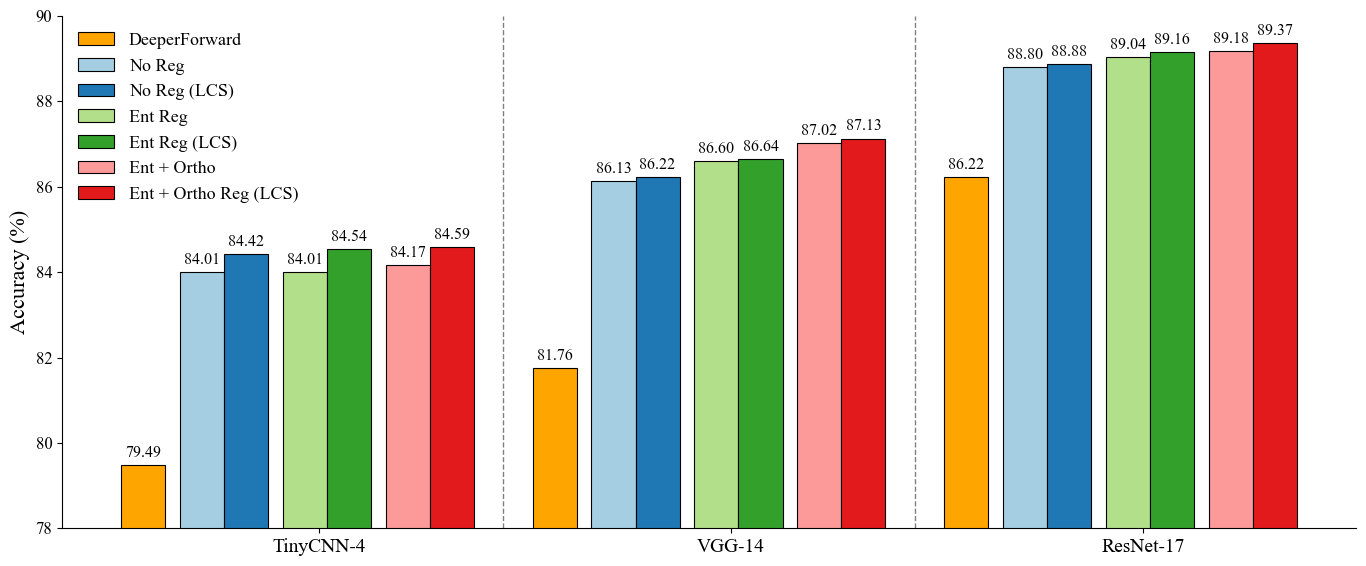}
% \vspace{1em}
\caption{Comparison of the effects of entropy (Ent) and orthogonality (Ortho) regularizations (Reg) applied to the learnable channel–class matrix in each CAW-Conv layer, and the influence of the layer contribution strategy (LCS) on final model accuracy on the CIFAR-10 dataset.}
\label{fig:3}
\end{figure}
%*****************************************

Table~\ref{table:results2} presents the classification performance of the proposed method on the more challenging CIFAR-100 and Tiny-ImageNet benchmarks and compares it against both conventional BP and DeeperForward~\cite{deeperff}.
On CIFAR-100, the proposed model achieves an accuracy of 63.52\%, improving upon the baseline DeeperForward model by nearly 10.5\%. Similarly, on Tiny-ImageNet, the proposed method attains 49.87\% accuracy, surpassing the reproduced DeeperForward result by approximately 8.5\%.
To further evaluate the scalability of the proposed approach, we increase the network capacity by tripling the number of convolutional channels (CH$\times3$). Under this setting, the proposed method achieves 69.74\% accuracy on CIFAR-100, representing a gain of more than 6\% over its standard configuration. In comparison, the corresponding DeeperForward-CH$\times$3 model reaches 60.28\%, which remains almost 9.5\% lower than the proposed approach. This result demonstrates that the proposed framework can utilize additional model capacity more effectively than existing FF-based methods. Although conventional BP still achieves the highest performance on both datasets, with accuracies of 70.70\% on CIFAR-100 and 61.8\% on Tiny-ImageNet~\cite{cifar100ref,tinyref}, the proposed method substantially narrows the gap between forward-only learning and gradient-based optimization.

%*****************************************
\begin{table}[t]
\centering
\caption{Accuracy comparison of the DeeperForward model and proposed method across varying convolutional channel sizes on the CIFAR-10 dataset. $\text{CH }\times (\alpha)$ indicates that the channel size is scaled by a factor of $\alpha$.}
\label{tab:channel}
\begin{tabular}{lccccc}
\toprule
\textbf{Architecture} 
& \textbf{DeeperForward} 
& \multicolumn{4}{c}{\textbf{Proposed Method}} \\
\cmidrule(lr){2-2} \cmidrule(lr){3-6}
& \textbf{CH$\times$(1.0)} 
& \textbf{CH$\times$(0.5)} & \textbf{CH$\times$(0.6)} & \textbf{CH$\times$(0.7)} & \textbf{CH$\times$(1.0)} \\
\midrule
TinyCNN-4  & 79.49 & 81.85 & 82.68 & 83.05 & 84.59 \\
VGG-14     & 81.26 & 80.98 & 82.53 & 84.38 & 86.91 \\
ResNet-17  & 86.22 & 85.73 & 86.52 & 87.51 & 89.37 \\
\bottomrule
\end{tabular}
\end{table}

%************************************
Figure~\ref{fig:3} investigates the individual and combined effects of the proposed entropy (Ent) and orthogonality (Ortho) regularization terms, as well as the layer contribution strategy (LCS), across three network architectures of increasing depth.
In the TinyCNN-4 setting, the baseline DeeperForward accuracy of 79.49\% is improved to 84.01\% even without any regularization, corresponding to a gain of almost 4.5\%. Similar improvements are observed for VGG-14 and ResNet-17, where the proposed model with no regularization exceeds DeeperForward by 4.37\% and 2.58\%, respectively.

The entropy regularization term further improves performance by encouraging balanced utilization of channels across different classes. Compared with the unregularized variant, applying entropy regularization consistently increases accuracy on deeper architectures. Although the improvements are moderate, their consistency suggests that entropy maximization prevents the channel-class assignment matrix from collapsing to a small subset of dominant channels, thereby promoting richer feature diversity.
The orthogonality regularization produces even larger improvements. When both entropy and orthogonality constraints are jointly applied, the accuracy increases to 84.17\%, 87.02\%, and 89.18\% for TinyCNN-4, VGG-14, and ResNet-17, respectively. Relative to the unregularized configuration, these gains correspond to improvements up to 0.9\%. The larger benefit obtained from orthogonality suggests that encouraging channel specialization plays a crucial role in enhancing discriminative feature learning.

An additional performance gain is achieved through the proposed layer contribution strategy. Under the entropy regularization setting, LCS provides accuracy improvements of up to 0.53\%, with the largest gain observed in the TinyCNN-4 architecture. These results demonstrate that weighting the contributions of different layers enables the model to more effectively exploit the complementary information captured at different representation levels, compared to the sub-layer selection mechanism employed in DeeperForward~\cite{deeperff}.
%************************************
\begin{figure}
\centering
\includegraphics[width=0.98\textwidth]{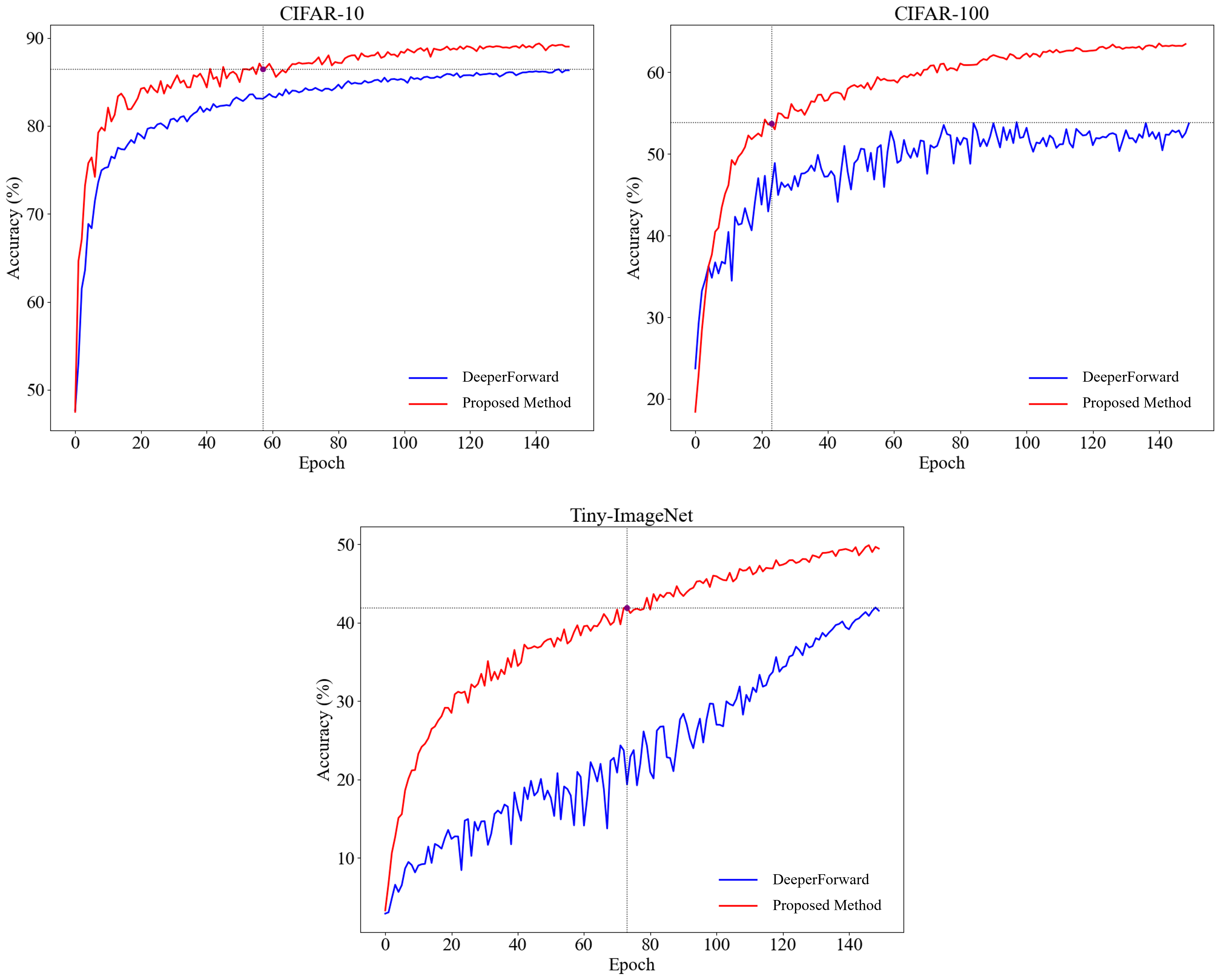}
% \vspace{1em}

\caption{Comparison of the rate of accuracy improvement between DeeperForward and our proposed method on test datasets of CIFAR-10, CIFAR-100, and Tiny-ImageNet. Proposed method's status at the global max accuracy of the DeeperForward method is indicated as a point.}
\label{fig:4}
\end{figure}

Table~\ref{tab:channel} demonstrates that the proposed method consistently outperforms the DeeperForward baseline while operating at reduced convolutional channel scales across different network architectures on the CIFAR-10 dataset. The notation of $\text{CH} \times (\alpha)$ denotes the use of a reduced channel size scaled proportionally by a factor of $\alpha$.

In the TinyCNN architecture with initial channel size $\mathrm{CH}$, DeeperForward achieves an accuracy of 79.49\%, whereas the proposed method attains 81.85\% with only $\mathrm{CH}\times(0.5)$, corresponding to a 50\% reduction in channel size, and further improves to 82.68\% and 83.05\% at $\mathrm{CH}\times(0.6)$ and $\mathrm{CH}\times(0.7)$, respectively, thereby surpassing the DeeperForward~\cite{deeperff} with substantially fewer channels. 

For the VGG architecture, the proposed method exceeds the DeeperForward~\cite{deeperff} accuracy of 81.26\% at the reduced scale of $\mathrm{CH}\times(0.6)$, where it achieves 82.53\%, and further widens the performance margin at $\mathrm{CH}\times(0.7)$ with an accuracy of 84.38\%, indicating that higher accuracy can be achieved without restoring full channel capacity. 

A similar trend is observed for ResNet, where DeeperForward reaches 86.22\% with full channels, while the proposed method already surpasses this performance at $\mathrm{CH}\times(0.6)$ with 86.52\% and at $\mathrm{CH}\times(0.7)$ with 87.51\%, showing that a 30\% to 40\% reduction in channel size is sufficient to outperform the baseline. 
The results indicate that proposed method, which utilizes learnable channel-class assignments, performs the task more efficiently than DeeperForward~\cite{deeperff}, which relies on static channel grouping.

Fig.~\ref{fig:4} compares the test accuracy of DeeperForward and proposed method on CIFAR-10, CIFAR-100, and Tiny-ImageNet over 150 epochs, revealing clear and consistent advantages of the proposed approach. Across both datasets, proposed method maintains higher accuracy throughout almost the entire training process, demonstrating superior learning efficiency and final generalization performance. 

On CIFAR-10, proposed method converges to around 89\%, whereas DeeperForward saturates at around 86\%. Also, the model trained with proposed method has achieved the best accuracy of the DeeperForward model at epoch 59.

On the more challenging CIFAR-100 task, the improvement becomes even more pronounced, with proposed method reaching approximately 64\% compared to DeeperForward's 52–54\%. Beyond final accuracy, proposed method shows significantly faster convergence: within the first 20 epochs, it consistently achieves noticeably higher accuracy, indicating more effective early-stage feature learning. The training dynamics also differ in stability, particularly on CIFAR-100, where DeeperForward exhibits large oscillations and irregular fluctuations. In contrast, proposed method maintains a smoother, more monotonic ascent, reflecting more stable optimization behavior. At the reference epochs marked by a vertical line, approximately at epoch 20, proposed method begins to outperform the DeeperForward method.

On the Tiny-ImageNet dataset, proposed method shows more consistent and faster improvements in accuracy, surpassing DeeperForward's performance by a significant margin after approximately 70 epochs. The accuracy of proposed method reaches approximately 50\% at the final epoch, while DeeperForward stabilizes at around 42\%. DeeperForward, by contrast, exhibits more erratic progress throughout training, with accuracy fluctuating and never achieving the steady upward trajectory of proposed method.

%%%%%%%%%%%%%%%%%%%%%%%%%%%%%%%%%%%%%%%%%%%%%%%%%%%%%

\section{Conclusion}\label{sec5}
In this work, we introduced a learnable channel-class assignment mechanism and a loss-aware layer contribution strategy to enhance the Forward-Forward algorithm within convolutional architectures. By enabling channels to dynamically specialize in class-specific representations, supported by entropy and orthogonality regularization, the proposed CAW-Conv layers significantly improve feature discriminability while reducing redundancy in representation learning. 
Compared to static channel grouping, the proposed CAW-Conv layers make more efficient use of channels through learnable channel-class assignments, achieving superior performance while reducing channel sizes by approximately 50\%. 
Furthermore, the adaptive layer contribution strategy leverages intermediate-layer predictions according to their loss values, weighting their contributions to the final prediction based on their local information level and learning progress. This mitigates the limitations of relying solely on final-layer outputs or selective sub-layer usage, which are common in similar forward-only learning methods.

Experiments across diverse benchmarks, including MNIST, Fashion-MNIST, CIFAR-10, CIFAR-100, and Tiny-ImageNet, demonstrate that proposed method consistently outperforms existing FF-based approaches, scales effectively to deeper networks, and narrows the performance gap with traditional backpropagation. These results confirm that integrating structural refinements, learnable channel specialization, and layer-wise credit assignment yields a more effective and high-performing forward-only learning framework.

Despite the fact that the learnable channel–class assignment matrix enhances performance compared to static channel grouping, the learning procedure of the proposed matrix can be further improved. Future research could focus on incorporating auxiliary mechanisms to more effectively guide the learning procedure of the proposed learnable matrix.
Furthermore, developing an effective mechanism for computing the global loss value and regulating its influence on the layer-wise local losses is another promising research direction that could improve the overall learning procedure.
%-----------------------------------------------------------

\printcredits

\bibliographystyle{cas-model2-names}

\bibliography{cas-refs}

\end{document}